\documentclass[10pt,twocolumn,letterpaper]{article}

\usepackage{cvpr}
\usepackage{times}
\usepackage{epsfig}
\usepackage{graphicx}
\usepackage{amsmath}
\usepackage{amssymb}
\usepackage{subfigure}
\usepackage{authblk}

\usepackage[breaklinks=true,bookmarks=false]{hyperref}

\cvprfinalcopy % *** Uncomment this line for the final submission

\def\cvprPaperID{****} % *** Enter the CVPR Paper ID here

\begin{document}

%%%%%%%%% TITLE
\title{Deep-SLAM++: Object-level RGBD SLAM based on\\ class-specific deep shape priors}

\author{Lan Hu}
\author{Wanting Xu}
\author{Haomin Shi}
\author{Kun Huang}
\author{Laurent Kneip}
\affil{ShanghaiTech University\authorcr   393 Middle Huaxia Road, Pudong, Shanghai, China \authorcr  \{hulan,xuwt,shihm,huangkun1,lkneip\}@shanghaitech.edu.cn}

\maketitle
\begin{abstract}
		%Recent years, the network for object generation has made a impressive progress. The main challenge to use the powerful ability of the network into reality is that the network is unstable sometime owing to  sensor noise, motion blur, data missing and view angle. In this work we propose a discrete seletion algorithm for the generated model according to the online measurements. Compared with the traditional reconstrcution method which fuse all the measurement together, we show that our algorithm which is much stable which independent of the pose estimation. More, we implement a object-level slam.
		In an effort to increase the capabilities of SLAM systems and produce object-level representations, the community increasingly investigates the imposition of higher-level priors into the estimation process. One such example is given by employing object detectors to load and register full CAD models. Our work extends this idea to environments with unknown objects and imposes object priors by employing modern class-specific neural networks to generate complete model geometry proposals. The difficulty of using such predictions in a real SLAM scenario is that the prediction performance depends on the view-point and measurement quality, with even small changes of the input data sometimes leading to a large variability in the network output. We propose a discrete selection strategy that finds the best among multiple proposals from different registered views by re-enforcing the agreement with the online depth measurements. The result is an effective object-level RGBD SLAM system that produces compact, high-fidelity, and dense 3D maps with semantic annotations. It outperforms traditional fusion strategies in terms of map completeness and resilience against degrading measurement quality.
	\end{abstract}
	
	%%%%%%%%% BODY TEXT
	\section{Introduction }
	
	The ability to perceive and localize within an unknown environment has long been recognized as a core driver of intelligent behavior in autonomous mobile systems, a problem commonly known as \textit{Simultaneous Localization And Mapping (SLAM)}. The focus of this work lies on indoor SLAM with a moving, consumer-grade RGBD camera, a problem to which we have seen a number of impressive solutions that confidently estimate both sensor trajectory as well as a relatively complete 3D model of the environment\cite{newcombe2011kinectfusion,kerl13iros,whelan2015elasticfusion}. Yet, the maps generated by such systems are purely geometric voxel or mesh-based surface representations that merely describe the boundaries of the environment. These representations can help to solve low-level problems such as path planning and navigation, but miss higher-level information such as the classes, shapes, and poses of objects relevant to the execution of more complex higher-level robotic tasks (e.g. find and collect all objects of a certain class).
	
	\begin{figure}[t]
		\centering
		\includegraphics[width=\columnwidth]{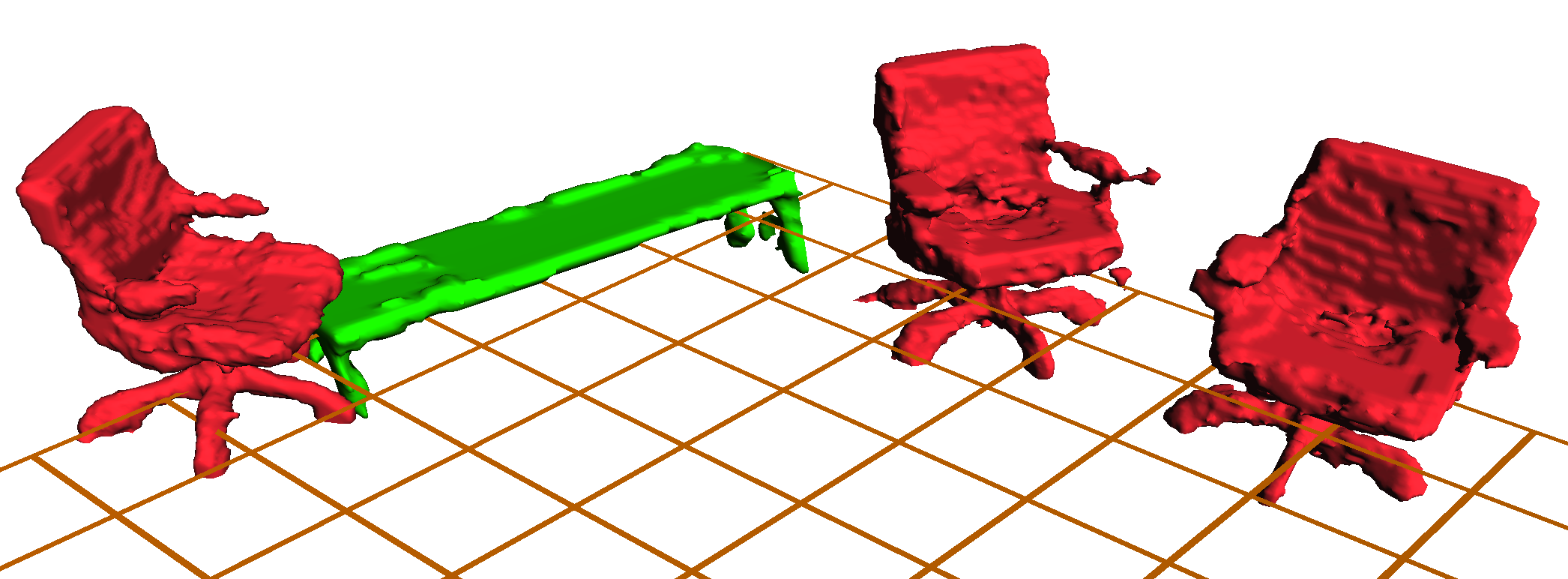}
		\caption{Example result created by Deep-SLAM++. Objects of different classes are represented by predictions from deep neural networks (red: chair, green: table).}
		\label{fig:rendering}
	\end{figure}
	
	In an effort to satisfy the higher-level perception needs of robots, the community has recently started to investigate SLAM systems that detect, segment, and reconstruct environments at the level of individual objects. A seminal contribution that generates such representations has been presented by Salas-Moreno et al.\cite{salas2013slam++}. They use an object detection mechanism to load full CAD models of object geometries which are then registered against the measurements taken by each view. While this already generates the desired output, it relies on the availability of exact object geometries. This contradicts the usual assumption in a SLAM scenario that the environment is not known in advance. Progress is given by the work of McCormack et al.\cite{mccormac2018fusion++}, who also segment out object instances in the input data, but reconstruct them individually using the online measurements and fused signed distance fields. However, while this enables operation in unknown environments, it does not employ any priors about object geometries despite the knowledge of each object class. As a result, the system fails to reconstruct sensible object geometries in the presence of occlusions and suffers from unmodeled measurement disturbances.
	
	In recent years, the ability to learn and apply sensible priors from large amounts of training data has been exhaustively demonstrated by the deep learning community. Starting from impressive results for image-based tasks such as object detection and pixel or instance-level semantic annotation, the most recent results even demonstrate full 3D object reconstruction and pose estimation from single views\cite{sun2018pix3d}. They dispose off the need for smoothness priors, and generate complete object geometries despite possible occlusions and missing data. However, while these results are very impressive, they often do not impose measurement consistency at all and---as a result---fail to provide confidence measures about the generated geometries. They are furthermore restricted to single objects seen from a limited number of nearby views, and regularly fail on real data with properties and artifacts that are insufficiently represented in the training data.
	
	Our contributions are as follows:
	\begin{itemize}
		\item Our primary result is a novel full-scale RGBD SLAM system that generates complete,  semantically annotated object-level representations of unknown indoor environments. It leverages object geometry priors for different classes given by modern neural networks as opposed to fusing all object measurements into a volumetric map. Lower level structures such as the ground plane are segmented and estimated in parallel.
		\item Network priors have varying performance owing to artifacts in the input data and the possibly unfavorable relative positioning between the camera and the object. Our proposed solution consists of incorporating a discrete selection mechanism into the estimation process, which picks the best among the multiple candidates generated in each view by evaluating depth measurement consistency.
		\item The resulting system represents an effective strategy to find a compromise between learned priors and online depth measurements, and the obtained object-level, joint geometric-semantic representations present a high level of completeness while having a low memory footprint.
	\end{itemize}
	
	Our SLAM framework can be divided into front-end and back-end modules. Section \ref{sec:front-end} presents the front-end modules, which covers relative camera pose estimation, object detection, segmentation, and object-to-camera alignment. It furthermore detects and fits the ground plane, and manages the addition of new objects or the correspondence establishment with existing ones. Section \ref{sec:back-end} then presents the back-end matters, which contains the graphical optimizer that refines absolute camera and object poses, and executes the discrete selection mechanism to identify those object geometry priors that maximize the agreement with the online measurements. It furthermore contains a loop closure module that effectively compensates for eventual drift accumulations. We conclude our exposition with our experimental results in Section \ref{rec:results}.
	
	\subsection{Review of existing works}
	
	\noindent\textbf{3D Model Generation:}
	Inferring the geometry of 3D object model from a single image is a difficult problem that requires both the recognition of the object or its class in the image, and the imposition of prior knowledge on its shape. Early attempts \cite{huang2015single} propose to borrow shape parts from existing CAD models. However, with the development of large-scale shape repositories like ShapeNet \cite{chang2015shapenet} and deep convolutional
	neural networks, researchers have built more scalable and efficient models in recent years \cite{wu20153d,girdhar2016learning,dai2017shape,yan2016perspective,yang20173d,hane2017hierarchical,novotny2017learning,wu2017marrnet}. While the majority of these approaches structure the object representation using a voxel grid, there have also been attempts to reconstruct objects with point clouds\cite{eigen2015predicting,groueix2018papier} or octave trees\cite{riegler2017octnet,tatarchenko2017octree,riegler2017octnetfusion}. Most of these works, however, focus on a single object seen from a limited number of views, and provide plain feed-forward predictions that do not evaluate the confidence in the generated model given more traditional measurement residuals (e.g. point-to-surface errors). The works of Zhu et al.\cite{zhu2017semantic} and Hu et al.\cite{hu2018dense} therefore embed the network decoder into a least squares residual minimization framework that iteratively refines the latent variables and camera poses. However, it is difficult to balance the influence of the network priors against the more traditional measurement residuals, and the approach remains restricted to a small set of nearby views observing a single object. Convergence furthermore strongly depends on the quality of the initial network prediction, which frequently returns unstable results depending on the camera viewing angle and the position of the object in the image. Our main insight is that---with the abundance of views generated in full SLAM scenarios---it is better to install a greedy search strategy in which we generate many 3D model predictions from many views, and use measurement fidelity to simply perform a discrete selection.
	
	%Both \cite{zhu2017semantic}\cite{hu2018dense}  try to use the measurement to refine the network generated objects, however, it is difficult to make a balance of the network and measurement.  However, we found that the factor influence the performance of the network is the input image, especially the view angle ans the position of the object in the input image. Instead to refining it with additional optimization with large computation. The greedy algorithm to choose the best generated model from all the different views. In our work, we use the network to generate the object and propose a algorithm to choose the best view for model.
	
	\noindent\textbf{Object-level SLAM:}
	Starting from simple geometric primitives such as points, planes, and lines, the community has moved on to the embedding of image-based semantic object detection modules into camera-based SLAM. Similar to the work of Salas-Moreno et al, Civera et al.\cite{civera2011towards} propose a semantic SLAM algorithm which extracts SURF features to recognize specific objects within a previously defined collection, and transfer the point cloud of a matched object as a rigid set of points into the map. Recent developments in deep learning have furthermore enabled the integration of rich semantic information into the estimation process. Mu et al. \cite{mu2016slam} also use an object detector to segment the depth measurements of an object. However, the work does not reconstruct full models of objects, but only estimates its object centers as a landmark in a sparse graph optimization framework. McCormac et al.\cite{mccormac2017semanticfusion} fuse predicted pixel-level semantic masks from a CNN in a dense, semantically annotated 3D map. In their second work \cite{mccormac2018fusion++}, they furthermore use the object segmentation to structure the 3D representation and install individual TSDF volumes for each object. The composition of the environment is hence reflected in both the semantic and the geometric forms of the representation. Further geometry-based semantic reconstruction methods have been presented \cite{sunderhauf2017meaningful,nakajima2019efficient,furrer2018incremental,pham2019real}, which all result in volumetric representations (or point clouds) with an object-level grouping. However, they only work well under the assumption of sufficiently accurate camera poses, and fail to impose any priors about object geometries despite knowledge of their class. CAD model based works \cite{salas2013slam++,avetisyan2018scan2cad} contrast with a bottom-up reconstruction of the scene geometry by consulting a shape database and aligning the retrieved model with the input scans. However, they only work if exact prior knowledge about the geometry of the objects in the environment is given in advance.
	To the best of our knowledge, we present the first full-scale object-level SLAM framework that successfully embeds 3D model generation networks into the estimation process, and applies it on real data.
	
	\subsection{Vocabulary and notations}
	To avoid confusion about detected objects in the image and object models generated in 3D, we henceforth denote the generated 3D object geometry priors as \textit{models}. For each frame $i$, we define the transformation matrix $\mathbf{T}_{wi}$ transforming points from the camera to the world frame. $\mathbf{T}_{wi}$ hence describes the absolute pose of each view. Accordingly, we define $\mathbf{T}_{ij}$ to be the relative Euclidean transformation from frame $j$ to frame $i$. We define  $\mathcal{O}$ to be the set of all models in an environment. The transformation from a camera to the model frame of a detected object is given by $\mathbf{T}_{oi}$, and the absolute pose of each object is described by $\mathbf{T}_{wo}$. All transformations are Euclidean rigid body transformations and represented by 4$\times$4 matrices containing a 3D position and a three DoF rotation matrix. The transformations may be parametrized minimially by a 6-vector $\mathbf{x}$ containing the 3D position $\mathbf{t}$ and a minimal parametrization of the rotation $\mathbf{R}$ (e.g. Rodriguez parameters). The observing frames of model $k \in \mathcal{O}$ are recorded in $\mathcal{L}_k$. The candidate models generated from each RGB frame in $\mathcal{L}_k$ are grouped in the set $\mathcal{C}_k$.
	
	\section{Front end}
	\label{sec:front-end}
\begin{figure}[b]
		\centering
		\includegraphics[width=\columnwidth]{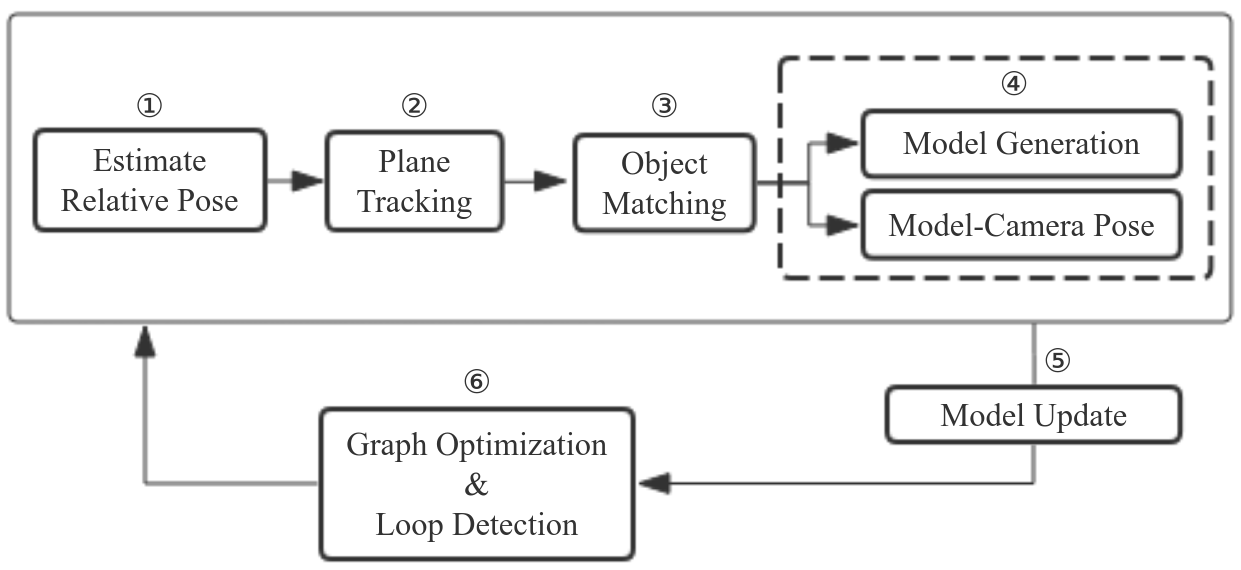}
		\caption{Overview of the modules and processing flow in our object-level SLAM framework. The modules in the grey-bounded box constitute the front-end and are executed for each new frame in order to set up the structure of a back-end optimization problem. For each new frame, we start by finding the relative pose $\mathbf{T}_{ij}$ with respect to the previous frame (\textcircled{\footnotesize{1}}). We then use $\mathbf{T}_{ij}$ to propagate the ground plane parameters into the current frame, resegment points on the plane, and optimize its relative location in the current frame (\textcircled{\footnotesize{2}}). We then detect and segment individual objects in the current frame, and try to match them with existing models (\textcircled{\footnotesize{3}}). If this succeeds, we calculate the camera to model pose, and if not, we generate a new model via CNN (\textcircled{\footnotesize{4}}). The back-end consists of a discrete part in which we select the best model among all candidates (\textcircled{\footnotesize{5}}), and a graphical optimizer that refines all camera and model poses (\textcircled{\footnotesize{6}}).}
		\label{fig:pipeline}
	\end{figure}

	An overview of all the modules in our pipeline is given in Figure \ref{fig:pipeline}. They can be grouped into a front-end that executes recurring tasks for each new frame and defines the structure of a back-end optimization problem, in which we then select good models and perform the actual refinement of absolute camera and model poses.
	The present section introduces the front-end modules, which contains a relative camera pose tracker, a plane segmentation and estimation module, an object detection, segmentation and matching module, a 3D model prior generation module, and a camera—to-model pose estimation module. The relative camera pose tracker relies on a standard Iterative Closest Point (ICP) algorithm with point-to-plane distance, and will not be discussed in further details. All other modules are introduced in the following.
	
	\subsection{Plane and object segmentation}
	\begin{figure}[t]
		\centering
		\includegraphics[width=\columnwidth]{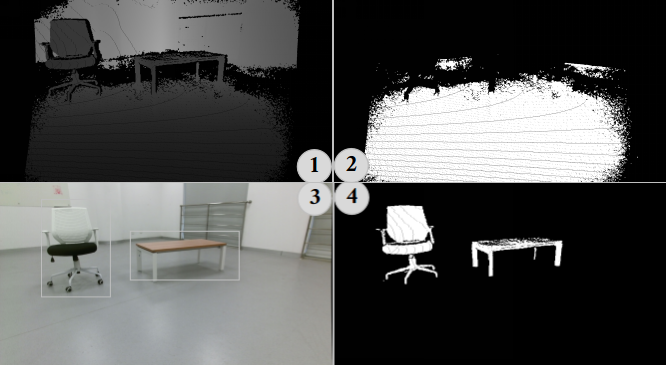}
		\caption{Example input images with segmented ground plane (top right), detected objects (bottom left), and segmented objects (bottom right).}
		\label{fig:plane_tracking}
	\end{figure}
	We initialize the ground plane by making a simple assumption about the initial camera viewpoint and the resulting relative location of the ground plane. We then perform dominant plane extraction within the limits of permissible plane orientations. In subsequent frames, the location of the ground plane is propagated by the tracked relative camera pose $\mathbf{T}_{ij}$, which supports the immediate re-identification of points on the ground. After optimizing the relative ground plane location using a robust error norm, we then apply the YOLO detector\cite{redmon2016you} to identify the bounding box of potential objects in the RGB frame. We then segment out precise object silhouettes by consulting the depth measurements within the bounding box, and using those that are close enough and not on the ground plane. Ignoring points on the ground is particularly useful for objects that are effectively placed on the ground. Object segmentation is used for later verification of model generations in the back-end, as well as for the extraction of relative object-to-camera transformations. Figure \ref{fig:plane_tracking} shows an example input RGBD frame with detected objects and the segmentation results.
	
	%Objects are very important semantic information, it is very difficult to get the exact silhouette of object, both the traditional object segmentation and learning-based object segmentation. We assume that all the objects are on the same ground plane,  once we know where the plane, the the object segmentation will be much easier. Fig.\ref{fig:plane_tracking} show the related image,  for every frames, we keep tracking the ground plane. And the measurements on the ground are the objects what we want.
	
	\subsection{Object matching}
	\label{sec:objMatch}
	%
	%How to incorporate all kind different representation into slam seamlessly is challengable. Since in SLAM, the main representation of map is TSDF and point cloud. The work \cite{zhu2017semantic} which use the sparse point cloud to representation, it could incorporate into the RGBD slam, but it could not provided dense representation of the object, since it is difficult to transform the point cloud into surface. However considering voxels , it is easily transformed into surface and also transform to pointcloud, so here we choose voxel as the representation of object. Recent work \cite{wu2017marrnet} have obtained the good performance, but its generated object are under camera view, which means that the object is not independent of the camera, is still like a point measurement. However in pix3d, the generated object is fixed orientation, which also obtained state-of-art results of the reconstruction. Referring to the pix3d, we train  our light network.
	By object matching, we understand the establishment of correspondences between 3D models and specific detected objects  in each frame. For each newly detected object, we need to decide whether it is a re-observation of an already existing object, or a new object. We first employ a voting strategy to re-identify models that occurred in any of the $5$ previous frames. To this end, we transform the depth points within each detected bounding box from the 5 previous frames into the current frame. We check if they end up inside a bounding box in the current frame, upon which we evaluate closest-point errors integrated over the entire transformed point set as a dissimilarity measure. A maximum error is applied in order to accept matching candidates. For each object detection in the current frame, we hence get a few, most likely redundant matched model candidates. We accept a correspondence if there is sufficient consensus among these candidates. If a detected object in the current frame can not be matched, we initialize a new model from it. Note that new models are only added if the bounding box is sufficiently spaced from the boundary of the image.
	
	\subsection{Generation of new models}
	Although the typical geometric representation of objects in SLAM is given by point clouds, meshes, or Truncated Signed Distance Fields (TSDF), we use binary occupancy grids as produced by the state-of-the-art model generation work of Sun et al. \cite{sun2018pix3d} (Pix3D). It proposes models from RGB only, which shows great benefits in situations of elevated distance to the object. Compared to the previous state-of-the-art given by Marrnet\cite{wu2017marrnet}---Pix3D generates models in a canonical fixed orientation along with an initial pose, which greatly simplifies the multi-view registration problem. The network is however not open-source, which is why we had to retrain our own. In order to accommodate for the real-time demands of SLAM systems, we furthermore reduced the input and output dimensions from 256$^{2}$ to 128$^{2}$ and from 128$^{3}$ to 64$^{3}$, respectively. This also meets the characteristics of realistic situations in which a single frame could contain multiple objects, and each object would assume a limited area in the image. Note that we use the full Pix3D network which includes depth and normal prediction, but input the pre-segmented objects from the RGB frame, which removes the perturbing influence of the background texture, and thus improves the performance of the overall model generation.
	
	\subsection{Camera-to-model pose estimation}
	\label{sec:modelpose}
	Camera-to-model pose estimation is cast as a point set registration problem, where one point set is obtained from the segmented object in the image, and the other one is obtained from the network output by simply replacing each occupied voxel with a point. Owing to the fixed resolution network output, it is clear that the latter is affected by an unknown scale. This unknown scale needs to be respected in the object-to-camera alignment step.
	
	We initialize the relative pose and scale with ordinary Procrustes analysis, which calculates the moments of the point cloud distribution in order to find the aligning transformation. However, to accommodate the fact that the measured point cloud only sees part of the model, we generate partial views of the generated model by a quantization of the model rotation about the vertical axis, and remove voxel centers which are occluded in a virtual orthographic object view with horizontal principal axis pointing at the model center. This works well under the assumption that the model has an upright placement on the ground plane. We analyze 8 rotation angles evenly distributed over the full 360$^{\circ}$, and perform ordinary Procrustes alignment with the measured depth points for each of the resulting partial views. The result showing the smallest point-to-point error is chosen as an initial relative camera-to-model pose.
	
	%We need to give a initialization of the rotation, observing that the object are orthogonal to the ground plane, which means  there only the rotation  is around $z$ axis. Simply, we split the $[0 \degree, 180 \degree]$ into $8$ bins $\Omega = \{i*\degree|i=0,1,\cdots 7\}$, then to minimize the error function
	%
	
	After the initial alignment, we refine the registration by applying ICP with a similarity transformation model and a point-to-plane error. The minimization objective is given as
	
	\footnotesize
	\begin{equation}
	\mathop{\text{argmin}}\limits_{\mathbf{x}, s} \sum \limits_{j}\|\left(\mathbf{R}(\mathbf{x})\mathbf{n}_j\right)^{T}\left(s\mathbf{R}(\mathbf{x})\mathbf{p}_j+ \mathbf{t}(\mathbf{x})-\eta(\mathbf{p}_j,\mathbf{x},s,\Omega)\right\|^2_2,
	\label{eq:p2planeICP}
	\end{equation}
	\normalsize
	where $\mathbf{x}$ denotes the parameters of $\mathbf{T}_{oc}$, $s$ the scale adjustment, $\mathbf{p}_j$ a point in the measured point cloud, $\mathbf{n}_j$ the normal vector at that point, and $\eta(\mathbf{p}_j,\mathbf{x},s,\Omega)$ a nearest neighbour function that returns $\mathbf{p}_j$'s nearest point in the model point set $\Omega$ given the current transformation parameters. Note that the alignment is done for both old and new models.
	
	\section{Back End} 
	\label{sec:back-end}
	The back-end consists of two parts. A graphical optimization framework in which all absolute poses are optimized in order to agree with the relative frame-to-frame and frame-to-model measurements. Note that we do not include all frames into the optimization, but only sufficiently spaced \textit{keyframes}. Besides the measurements, also the structure of the graph is defined by the correspondence-establishment in the front end. The second part consists of updating the model and also the dependent camera-to-model transformations  depending on the level of agreement with the depth measurements. We start with the latter part.
	\subsection{Model update}
	\begin{figure}[b]
		\centering
		\includegraphics[width=0.9\columnwidth]{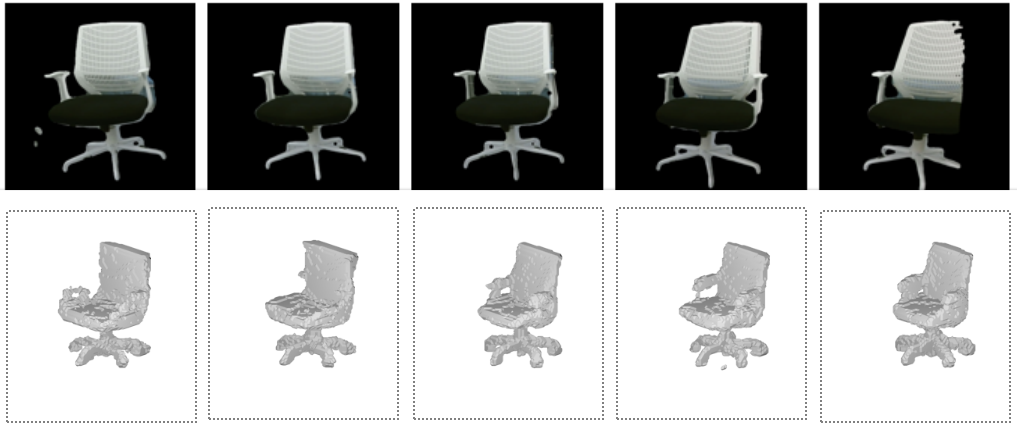}
		\caption{The sensitivity of the network's model generation. The first row shows inputs with marginal differences, and the second row illustrates the potentially large variations in the generated models.}
		\label{fig:network1}
	\end{figure}
	As illustrated in Figure \ref{fig:network1}, the quality of the network prediction typically varies strongly as the input is changed. In the worst case, even small changes in the viewing angle, the distance to the object, and the position of the object may lead to uncontrolled behavior in the network output, which is a common problem of network predictions going hand in hand with missing confidence measures. Our solution to produce more confident predictions consists of generating multiple proposals  $\mathcal{C}_k$  from distinct views $\mathcal{L}_k$, and performing a discrete selection by consulting the agreement with an accumulation of the actual depth measurements of model $k$.
	
	The main problem is that a bad model generation can have a negative impact on the model-to-camera transformations, and the latter are needed in order to register all measurements in the model frame and cross-check the quality of the generation. We could alternatively perform the registration by using direct relative frame-to-frame transformations obtained from ICP, but they often prove to be of insufficient accuracy as well. We solve this by a greedy strategy in which all camera-to-model transformations are derived for each model candidate, individually. In this way, the quality of the measurement registration depends solely on the quality of the currently verified model candidate. If one of the candidates has good quality, the camera-to-model poses will have good accuracy, and the result will be a high agreement between model and measurements.

	\subsection{Graph optimization}
	
	The remaining objective of our framework is the recovery of refined absolute camera and model poses, denoted $\mathbf{T}_{wi}$ and $\mathbf{T}_{wo}$, respectively. We formulate this problem as a graphical optimization problem in which correspondences between neighbouring keyframes or keyframes and models are reflected as edges between nodes, the latter representing the absolute pose variables $\mathbf{x}_{wi}$ and $\mathbf{x}_{wo}$ to be optimized. For each edge, we then have a measurement obtained from the point-to-plane based ICP objectives for direct frame-to-frame tracking and model-to-camera pose estimation (cf. Section \ref{sec:modelpose}). We denote these measurements $\mathbf{z}_{ij}$ and $\mathbf{z}_{io}$. We furthermore define the information matrices $\mathbf{\Omega}_{ij}$ and $\mathbf{\Omega}_{io}$ to describe the uncertainty of the calculated relative poses. We obtain them via Hessian approximations given by $\mathbf{H}=\mathbf{J}^T\cdot\mathbf{J}$, where $\mathbf{J}$ represents the Jacobian of the point-to-plane residuals evaluated at the final solution returned by ICP. Note that the correspondences remain fixed during the extraction of the Jacobians.
	
	%$\mathbf{Z}_{i,o_j}$ (dotted light green) denotes
	%the 6DoF measurement of object $j$ in frame $i$ and $\Sigma_{i,o_j}^{-1}$ its
	%Information can be estimated using the approximated Hessian $\Sigma_{i,o_j}^{-1} = J^T*J$ (with $J$ being the Jacobian from the final iteration of the object ICP). 
	%$\mathbf{Z}_{i,i+1}$ (dotted blue) is the relative ICP constraint between camera $i$ and
	%$i+1$, with $\Sigma_{i,i+1}^{-1}$ the corresponding inverse covariance. The
	%absolute poses $\mathbf{T}_wi$ and $\mathbf{T}_{w,o_j}$
	%are variables which are modified during the optimisation, while $\mathbf{Z}_{i,oj}$
	%and $\mathbf{Z}_{i,i+1}$ are measurements and therefore constants. All variables and
	%measurements have 6DoF and are represented as members of $SO(3)$.
	
	The residual errors at each edge are finally defined as
	\begin{equation}
	\left\{\begin{matrix}
	\mathbf{r}_{io}=\operatorname{t2v}\left(\mathbf{T}^{-1}(\mathbf{z}_{io})\cdot\mathbf{T}^{-1}(\mathbf{x}_{wi})\mathbf{T}(\mathbf{x}_{wo})\right) \\
	
	\ \ \mathbf{r}_{ij}=\operatorname{t2v}\left(\mathbf{T}^{-1}(\mathbf{z}_{ij})\cdot\mathbf{T}^{-1}(\mathbf{x}_{wi})\mathbf{T}(\mathbf{x}_{wj})\right),
	\end{matrix}\right.
	\end{equation}
	where $\operatorname{t2v}$ denotes the function that extracts the minimal 6-vector parametrization from a 4$\times$4 transformation matrix.
	
	As our framework is also able to measure and map the ground plane, it raises the question of how to constrain the corresponding parameters. We parametrize the ground plane measurement in each frame by a vector $\mathbf{p}_i=\left[\begin{matrix}\mathbf{n}_i^T & d_i\end{matrix}\right]^T$, where $\mathbf{n}_i$ represents the normal vector of the plane, and $|d_i|$ its orthogonal distance from the origin. We furthermore introduce the global ground plane coordinates $\mathbf{p}=\left[\begin{matrix}\mathbf{n}^{T} & d\end{matrix}\right]^{T}$ as a further optimization variable. Each frame in which the ground plane is visible will give rise to another residual error, which is given by
	\begin{equation}
	\mathbf{r}_{pi} = \mathbf{p}-\left[\begin{matrix}\mathbf{R}(\mathbf{x}_{wi})\mathbf{n}_i\\d_i-(\mathbf{R}(\mathbf{x}_{wi})\mathbf{n}_i)^{T}\mathbf{t}_{wi}\end{matrix}\right],
	\end{equation}
	Parametrizing the ground plane lets us furthermore constrain the orientation of upright objects to align with the vertical direction. This introduces yet further residuals of the form
	\begin{equation}
	\mathbf{r}_{po}=\mathbf{n}-\mathbf{R}(\mathbf{x}_{wo})\mathbf{v},
	\end{equation}
	where $\mathbf{v}$ is simply the vector $(0,0,1)^{T}$. In this work, we restrict ourselves to normal scenarios in which all objects are standing upright on the horizontal plane. Note that there is a simple unit-norm side constraint on $\mathbf{n}$, which can again be enforced implicitly by choosing a minimal 2-parameter parametrization for the vertical direction $\mathbf{n}$. Also note that there are further information matrices $\mathbf{\Omega}_{\mathbf{p}_i}$, which can be extracted for each ground plane measurement in each frame by again approximating the Hessian at the minimum of the robust plane fitting objective.
	\begin{figure}
		\centering
		\includegraphics[width=0.96\columnwidth]{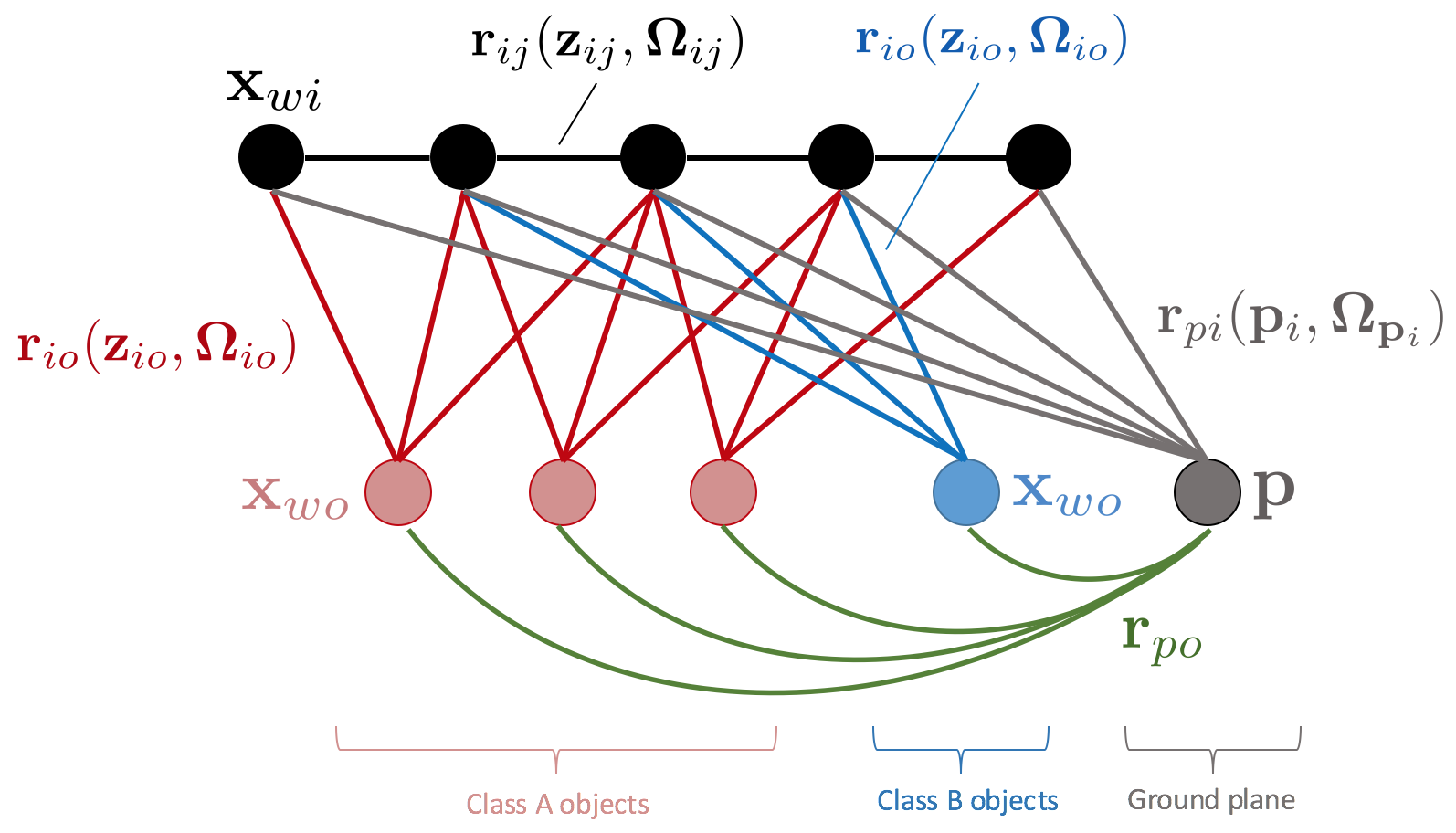}
		\caption{The graphical model of our back-end optimization framework with five keyframes, multiple objects of two classes, and the ground plane.}
		\label{fig:graph}
	\end{figure}
	The resulting graph optimization problem is depicted in Figure \ref{fig:graph}. It contains edges along which residuals of the above four types are being minimized. Note that the nodes are grouped by object classes, and that the ground plane node is connected to all other frames and objects. The graph optimizer finally minimizes the energy
	\begin{equation}
	\begin{split}
	E = &
	\sum_{\mathbf{z}_{io}}
	\mathbf{r}_{io}^{T}\mathbf{\Omega}\mathbf{r}_{io}
	+ \lambda_1 \cdot \sum_{\mathbf{z}_{ij}}
	\mathbf{r}_{ij}^{T}\mathbf{\Omega}\mathbf{r}_{ij}\\
	& + \lambda_2 \cdot \sum_{i} \mathbf{r}_{pi}^{T} \mathbf{\Omega}_{\mathbf{p}_i} \mathbf{r}_{pi}
	+ \lambda_3 \cdot \sum_{o} \mathbf{r}_{po}^{T} \mathbf{r}_{po},
	\end{split}
	\end{equation}
	where the information matrices are used to perform statistical reweighting of the errors, and the factors $\lambda_1$, $\lambda_2$ and $\lambda_3$ are adjusting the general trade-off between the different parts of the optimization objective. The resulting generalised least-squares problem is solved using Levenberg-Marquardt.

	\subsection{Loop detection}
	Loop closure in SLAM occurs when---after a certain period of time---a location is revisited, and the recognition of the previously visited place is used to compensate for drift accumulation along the loop. In our scenario, drift and the absence of a dedicated loop detection mechanism would lead to a failure to reidentify existing objects, and hence to redundant model definitions. Small loops lead to only small drift accumulations, which may still permit the identification of true correspondences between earlier frames and objects using the method outlined in Section \ref{sec:objMatch}. For larger loops, however, a dedicate loop detection mechanism is needed. We propose to rely on the semantic landmarks, and loop detection triggers as soon as a similar constellation of at least two models is repeatedly observed. The similarity of two constellations is evaluated by first checking class consistency for the objects in question, followed by a check of the similarity of the relative poses between 3D models.
	
	%But in our work,  it also correct wrong generation of the repeated model  which caused by  the object matching failure due to the drift. For ICP-based registration methods,generally small loops are regularly closed using the standard ICP tracking mechanism. Larger loop closures, where the drift is too much to enable matching via predictive ICP,  are detected  using the local graph a module on based matching fragments within the main long-term graph in the same manner as in relocalization. We use the observed objects as feature to do the loop detection, only when at least two objects are registered well, then we decide there is a loop closure.
	
	\section{Results}
	\label{rec:results}
	
\begin{figure}[t]
		\centering
		\includegraphics[width=\columnwidth]{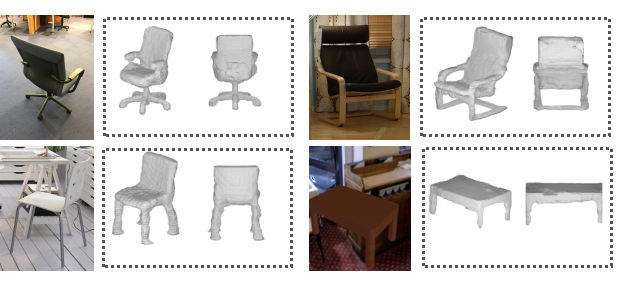}
		\caption{Example reconstructions of chairs and tables using our custom Pix3D networks.}
		\label{fig:network}
	\end{figure}

	We perform similar experiments than the seminal work of \cite{salas2013slam++}, and obtain results on datasets taken with a Kinect camera in regular indoor office environments with chairs and tables distributed randomly over the ground plane. The two object classes are detected by YOLO, and we furthermore train one Pix3D network per object class in order to generate model geometries. The performance of our networks is carefully evaluated on large test sets before the actual application in our SLAM framework. Example reconstructions of chairs and tables from the test set are indicated in Figure \ref{fig:network}. Apart from a difference in the input and output resolution, our network has similar performance than the original Pix3D network. For further details, we therefore kindly invite the reader to look up the work of Sun et al.\cite{sun2018pix3d}. The present results are focused on the embedding of the network into our proposed Deep-SLAM++ framework.
	
	%We use the kinect to collect the data in two small rooms which contains 18 chair, the other contains 9 chairs and one table, Fig \ref{fig:graph_map} shows the structure of the first room. The size of the image is $1920*1080$.
	%
	
	\subsection{Overall performance}
	
	\begin{figure}[b]
		\centering 
		\subfigure{ 
			\includegraphics[width=0.45\columnwidth]{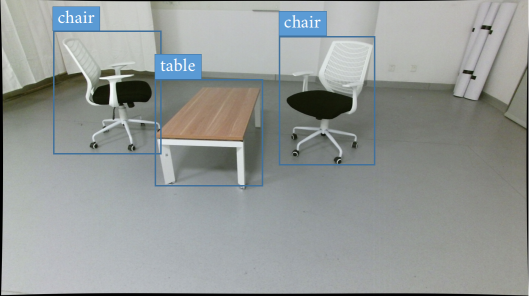} }
		\subfigure{ 
			%\label{fig:subfig:b} %% label for second subfigure 
			\includegraphics[width=0.45\columnwidth]{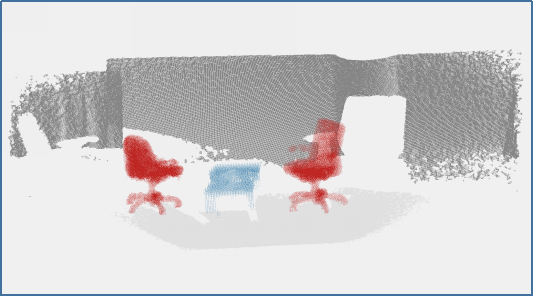} }
		\caption{Example of the scene reconstruction, the left figure shows the RGB image with object detection results, the right one illustrates the point representation of the scene. The chairs (in red) and the table (in blue) are generated point clouds rather than original measurements.} 
		\label{fig:scene_reconstruction} %% label for entire figure 
	\end{figure}
	
	\begin{figure*}[t]
		\includegraphics[width=\textwidth,height=4cm]{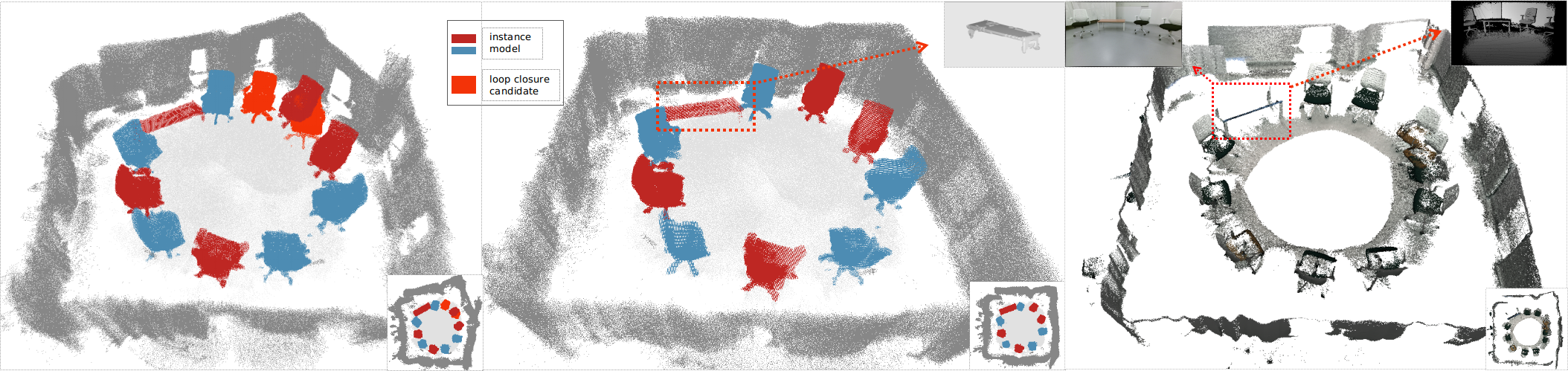}
		\caption{Left: Semantically annotated Deep-SLAM++ result before loop closure, the two chairs in brighter red are the loop detection candidates. Center: The result after final graph optimization with loop closure. Right: Result obtained by ElasticFusion, which is affected by many gaps in the reconstruction. The red bounding-box highlights the contrast in the completeness of the table reconstruction for which only a few depth measurements are available. Note that the object colors are not indicating object class, but simply alternating between blue and red.}
		\label{fig:map}
	\end{figure*}
	
	We test our framework on several indoor sequences of varying scale, all captured with a Kinect V2 sensor. Figure \ref{fig:scene_reconstruction} shows a first example, where the camera is moved in front of a regular scene with two chairs and one table. The left image shows the object detections returned by YOLO, and the right image shows the result of the aligned depth frames produced by our SLAM framework. Note that the points on the chairs (in red) and the table (in blue) are not the original points measured by the camera, but the---generally more complete---output generated by the networks.
	
	Figure \ref{fig:map} shows our main result, which is a comparison against the popular open-source RGBD-SLAM framework \textit{ElasticFusion}\cite{whelan2015elasticfusion} on a slightly larger sequence. The left and center image show our semantic reconstruction before and after loop closure, where the segmented points on the objects have again been replaced by the model points generated by our class-specific networks. The right image shows the reconstruction obtained by ElasticFusion. As can be observed, owing to reflections and inclined viewing angles, the reconstruction obtained by ElasticFusion contains many gaps and incomplete object geometries. Deep-SLAM++ has the advantage of only requiring RGB information for generating model geometry proposals, and as a result provides complete reconstructions in which even occluded parts of the structure are contained. Though Deep-SLAM++ also needs depth measurements to discriminate model candidates, it proves the ability to do this with a very limited amount of correct depth measurements. Taking the table in the scene as an example, the result from ElasticFusion suggests that there are poor depth measurements on the surface of the table. The measurements of a few edges of the table are sufficient to generate a complete and sensible object geometry.
	%Note however that the number and quality of the geometry proposals still has a dependency on the depth information, as we maintain a hard requirement for model geometry proposals to be verified via correct depth measurements.
	
	\begin{figure}[b]
		\centering
		\includegraphics[width=0.9\columnwidth]{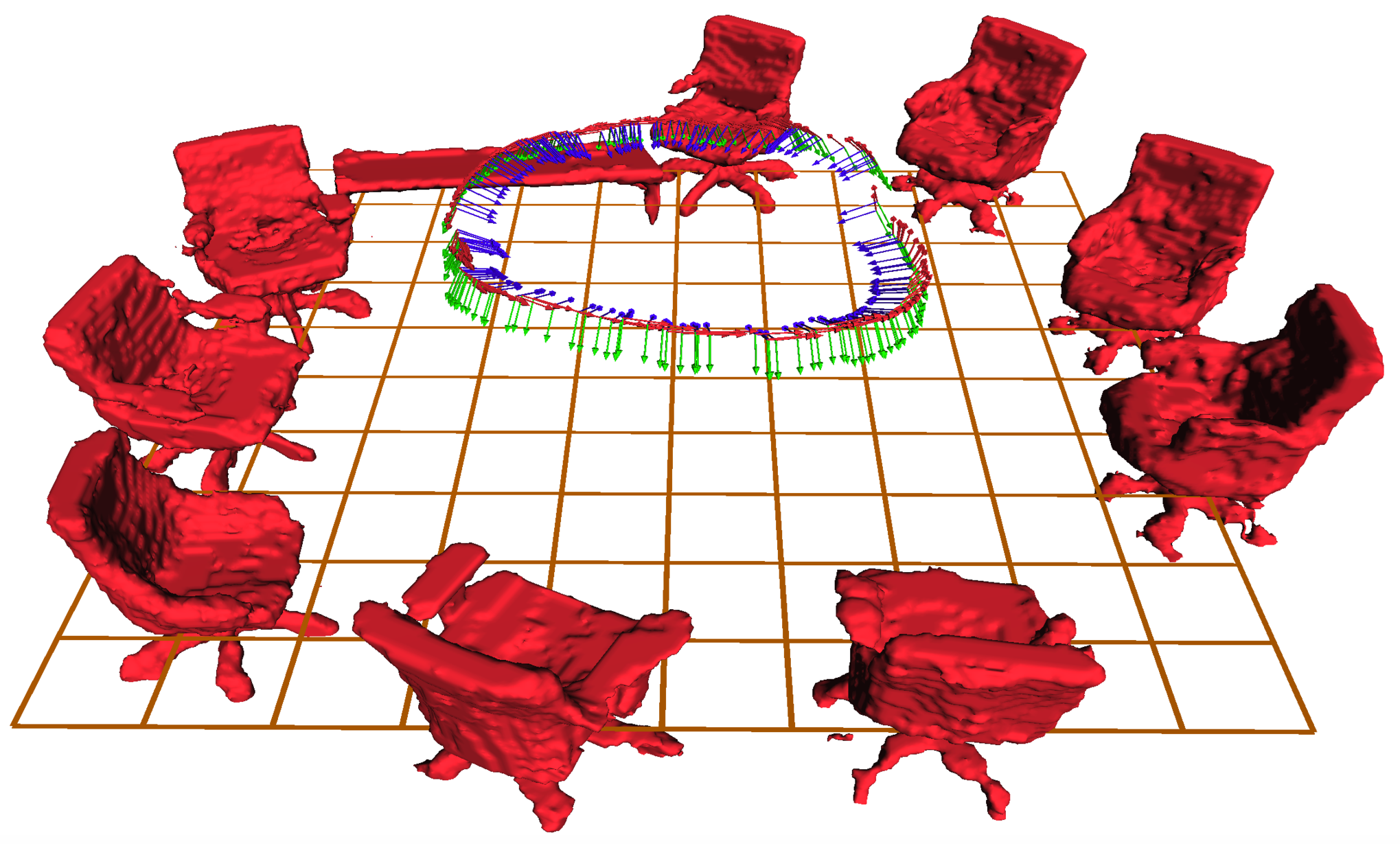}
		\caption{3D rendering of the generated representation of the environment.}
		\label{fig:rendering2}
	\end{figure}
	
	Our comparison remains qualitative, as our aim is not to outperform in terms of localization performance. Our aim is to demonstrate how the 3D scene representations produced by our framework are more complete and contrast with lower-level pipelines by containing the desired higher-level properties such as object-level segmentation and object poses, geometries, and classes. Figure \ref{fig:rendering2} shows a rendering of the internal representation given by all object models and graph optimization variables at the end of the loop. Since each object can now be represented by the 100 latent variables in the network only, and the points on the ground can be replaced by the ground plane parameters, the size of the entire map is reduced by at least two orders of magnitude compared to the point cloud size returned by ElasticFusion. A video of the execution is contained in the supplementary material.
	
	\subsection{Incremental quality of the 3D models}
\begin{figure}[t]
		\centering
		\includegraphics[width=0.4\textwidth]{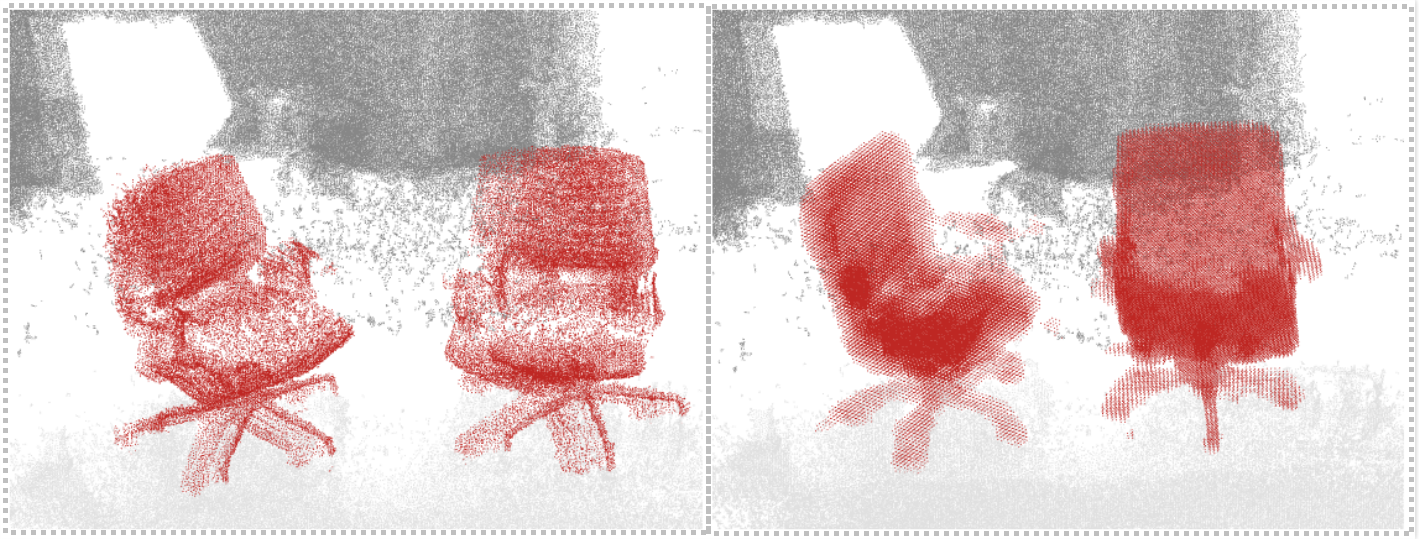}
		\caption{The figure illustrates the advantage of the network generated model. The left figure shows the fused measurements from $15$ continuous frames, and the blur in the alignment. The right figure illustrates the proposal of the network, which shows a significantly reduced dependency on such depth measurement perturbations.}
		\label{fig:two_chair}
	\end{figure}

\begin{figure}[b]
		\centering
		\includegraphics[width=0.4\textwidth]{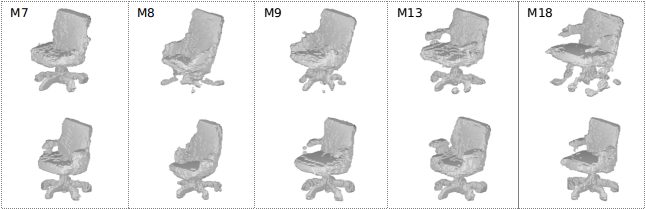}
		\caption{Illustration of the model improvement. Each column contains different versions of the same object. The first row indicates the initial proposal, while the second row indicates the replacing improved model generated from a subsequent view.}
		\label{fig:two_update}
	\end{figure}

	Figure \ref{fig:two_chair} contains fused depth points from 15 nearby views. As can be clearly observed, the depth points do not result in a crisp structure of the object but present a significant amount of blur. While this is partially related to inaccurate camera poses, it is also related to unmodeled effects in the measurements such as mixed pixels, reflections, missing data, or even biases in the measurements depending on the surface properties and the angle of inclination. In ElasticFusion, the blur in the alignment translates into 3D points with low support. As a result, they are ignored in the final output which further sparsifies the representation of the environment.
	Our model generation improves resilience with respect to such unmodeled effects. The right image in Figure \ref{fig:two_chair} shows the result of replacing all points on the chair by our model generation. It can be observed that the structure becomes clearer even within the limits of the network's output resolution. The performance of the network still has quite large variability, and does not always return good proposals at its first attempt. In order to incrementally refine the models, we check its alignment with the depth measurements and replace a model if a better one is found through a prediction from a subsequent view. Figure \ref{fig:two_update} shows the incremental improvement of different objects along the loop. Note that the chairs are all of the same type, which demonstrates high repeatability in this model update procedure. 
	
	We also compare the Intersection over Union (IoU) and Chamfer distance (CD)\cite{sun2018pix3d} between the generated model and a reference. In order to demonstrate the improvement, we the distances are evaluated for both the initial and the improved models. Since there is no ground truth model, we pick a similar one from the ShapeNet. As can be observed in Table \ref{tab:error1}, both the IoU and the CD error verify that our DeepSLAM++ selects improved models from subsequent views.
	\begin{table}
		\centering
		\begin{tabular}{|c|c|c|c|}\hline
			& IoU & CD \\\hline
			Initialized model   & 0.2068 & 0.0947 \\
			Improved model   & \textbf{0.2837}&\textbf{0.0812}\\\hline
		\end{tabular}
		\caption{Average IoU  and CD errors of initial and improved reconstructed 3D models.}
		\label{tab:error1}
	\end{table}
	%\begin{figure}
	%    \centering
	%    \includegraphics[width=0.2\textwidth]{render2.png}
	%    \caption{ There are two model of each rows, which has been update two times}
	%    \label{fig:two_update}
	%\end{figure}.
	%-----------------------------

	\subsection{Loop closure }
\begin{figure} 
		\centering 
		\includegraphics[width=\columnwidth]{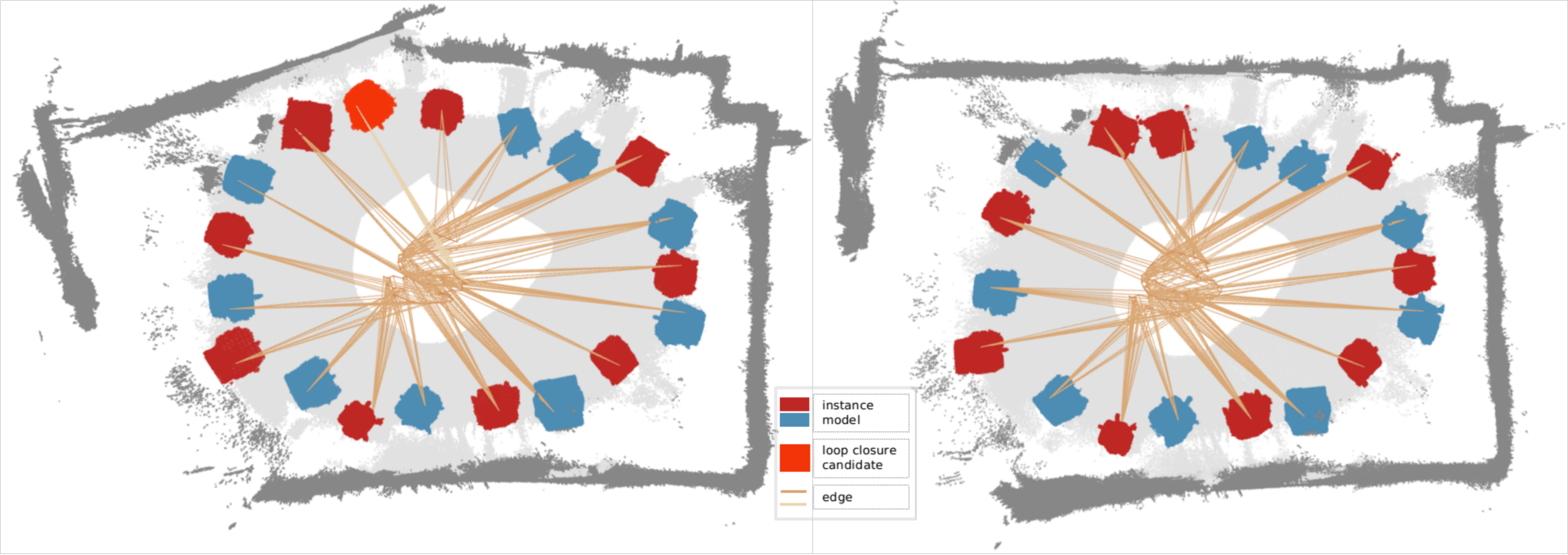}
		\caption{Left: Environment representation before the loop closure. The bright red chair is the loop closure candidate, which has been identified as having a similar relative placement to at least one neighbour of a certain class than a previous object of the same class. Right: Result after loop closure correction. Note that all objects are correctly identified as chairs, and the alternation between red and blue is just a visualization artefact. The figures furthermore encode the graph connections between frames and objects, as well as points identified as lying on the ground plane (light grey).}
		\label{fig:graph_map} %% label for entire figure 
	\end{figure}

	Figure \ref{fig:map} already illustrates our framework's ability to detect loops and close them through our back-end graph optimization. Loops are detected by considering the similarity in the classes and relative arrangement of a group of at least 2 nearby objects in the scene. Active detection of loops is important in order to remove drift and prevent redundant object placements. Figure \ref{fig:graph_map} shows our final experiment, in which we execute a longer trajectory in a larger room. Owing to the increased distance to the background, the quality of the relative pose estimates is degrading, which in turn leads to a more important drift accumulation along the loop. The bright red chair again illustrates the loop closure candidate, which has been identified as having a similar relative placement to other objects. As can be observed, the drift is effectively removed after the loop closure optimization.
	
	% \twocolumn[{%
	% \renewcommand\twocolumn[1][]{#1}%
	% \begin{center}
	%     \centering
	%     \includegraphics[width=\textwidth]{map2.png}
	%     \caption{The pipeline of our work.}
	%     \label{fig:pipeline}
	% \end{center}%
	% }]
	
	%----------------------
	
	\section{Discussion}
	We present the first room-scale object-level RGBD SLAM system that successfully embeds object geometry priors given by a deep neural network into the estimation process. The varying performance of the network is encountered by a greedy strategy that selects the best of many priors in terms of its agreement with the depth measurements. Rather than ignoring the depth measurements altogether, this paradigm realizes an interesting trade-off between prior knowledge and online measurements, where the latter are used to evaluate the confidence of each prediction. It efficiently generates sensible, complete object geometries, and therefore contrasts with more computationally demanding bottom-up object reconstructions from online measurements only. While the datasets in this work are designed to clearly demonstrate the feasibility of this approach on real data, they capture well structured scenes with a limited number of object classes and occlusions, and a limited degree of object stapling and clutter. Our current efforts are directed at ensuring correct operation in more complicated scenes in which these conditions are less fulfilled.
	
{\small
\bibliographystyle{ieee}
\bibliography{ms}

\begin{thebibliography}{10}\itemsep=-1pt

\bibitem{avetisyan2018scan2cad}
A.~Avetisyan, M.~Dahnert, A.~Dai, M.~Savva, A.~X. Chang, and M.~Nie{\ss}ner.
\newblock Scan2cad: Learning cad model alignment in rgb-d scans.
\newblock {\em arXiv preprint arXiv:1811.11187}, 2018.

\bibitem{chang2015shapenet}
A.~X. Chang, T.~Funkhouser, L.~Guibas, P.~Hanrahan, Q.~Huang, Z.~Li,
  S.~Savarese, M.~Savva, S.~Song, H.~Su, et~al.
\newblock Shapenet: An information-rich 3d model repository.
\newblock {\em arXiv preprint arXiv:1512.03012}, 2015.

\bibitem{civera2011towards}
J.~Civera, D.~G{\'a}lvez-L{\'o}pez, L.~Riazuelo, J.~D. Tard{\'o}s, and
  J.~Montiel.
\newblock Towards semantic slam using a monocular camera.
\newblock In {\em Intelligent Robots and Systems (IROS), 2011 IEEE/RSJ
  International Conference on}, pages 1277--1284. IEEE, 2011.

\bibitem{dai2017shape}
A.~Dai, C.~R. Qi, and M.~Nie{\ss}ner.
\newblock Shape completion using 3d-encoder-predictor cnns and shape synthesis.
\newblock In {\em Proc. IEEE Conf. on Computer Vision and Pattern Recognition
  (CVPR)}, volume~3, 2017.

\bibitem{eigen2015predicting}
D.~Eigen and R.~Fergus.
\newblock Predicting depth, surface normals and semantic labels with a common
  multi-scale convolutional architecture.
\newblock In {\em Proceedings of the IEEE international conference on computer
  vision}, pages 2650--2658, 2015.

\bibitem{furrer2018incremental}
F.~Furrer, T.~Novkovic, M.~Fehr, A.~Gawel, M.~Grinvald, T.~Sattler,
  R.~Siegwart, and J.~Nieto.
\newblock Incremental object database: Building 3d models from multiple partial
  observations.
\newblock In {\em 2018 IEEE/RSJ International Conference on Intelligent Robots
  and Systems (IROS)}, pages 6835--6842. IEEE, 2018.

\bibitem{girdhar2016learning}
R.~Girdhar, D.~F. Fouhey, M.~Rodriguez, and A.~Gupta.
\newblock Learning a predictable and generative vector representation for
  objects.
\newblock In {\em European Conference on Computer Vision}, pages 484--499.
  Springer, 2016.

\bibitem{groueix2018papier}
T.~Groueix, M.~Fisher, V.~G. Kim, B.~C. Russell, and M.~Aubry.
\newblock A papier-m{\^a}ch{\'e} approach to learning 3d surface generation.
\newblock In {\em Proceedings of the IEEE conference on computer vision and
  pattern recognition}, pages 216--224, 2018.

\bibitem{hane2017hierarchical}
C.~H{\"a}ne, S.~Tulsiani, and J.~Malik.
\newblock Hierarchical surface prediction for 3d object reconstruction.
\newblock In {\em 2017 International Conference on 3D Vision (3DV)}, pages
  412--420. IEEE, 2017.

\bibitem{hu2018dense}
L.~Hu, Y.~Cao, P.~Wu, and L.~Kneip.
\newblock Dense object reconstruction from rgbd images with embedded deep shape
  representations.
\newblock {\em arXiv preprint arXiv:1810.04891}, 2018.

\bibitem{huang2015single}
Q.~Huang, H.~Wang, and V.~Koltun.
\newblock Single-view reconstruction via joint analysis of image and shape
  collections.
\newblock {\em ACM Transactions on Graphics (TOG)}, 34(4):87, 2015.

\bibitem{kerl13iros}
C.~Kerl, J.~Sturm, and D.~Cremers.
\newblock Dense visual {SLAM} for {RGB-D} cameras.
\newblock In {\em IROS}, Tokyo, Japan, 2013.

\bibitem{mccormac2018fusion++}
J.~McCormac, R.~Clark, M.~Bloesch, A.~Davison, and S.~Leutenegger.
\newblock Fusion++: Volumetric object-level slam.
\newblock In {\em 2018 International Conference on 3D Vision (3DV)}, pages
  32--41. IEEE, 2018.

\bibitem{mccormac2017semanticfusion}
J.~McCormac, A.~Handa, A.~Davison, and S.~Leutenegger.
\newblock Semanticfusion: Dense 3d semantic mapping with convolutional neural
  networks.
\newblock In {\em Robotics and Automation (ICRA), 2017 IEEE International
  Conference on}, pages 4628--4635. IEEE, 2017.

\bibitem{mu2016slam}
B.~Mu, S.-Y. Liu, L.~Paull, J.~Leonard, and J.~P. How.
\newblock Slam with objects using a nonparametric pose graph.
\newblock In {\em Intelligent Robots and Systems (IROS), 2016 IEEE/RSJ
  International Conference on}, pages 4602--4609. IEEE, 2016.

\bibitem{nakajima2019efficient}
Y.~Nakajima and H.~Saito.
\newblock Efficient object-oriented semantic mapping with object detector.
\newblock {\em IEEE Access}, 7:3206--3213, 2019.

\bibitem{newcombe2011kinectfusion}
R.~A. Newcombe, S.~Izadi, O.~Hilliges, D.~Molyneaux, D.~Kim, A.~J. Davison,
  P.~Kohi, J.~Shotton, S.~Hodges, and A.~Fitzgibbon.
\newblock Kinectfusion: Real-time dense surface mapping and tracking.
\newblock In {\em Mixed and augmented reality (ISMAR), 10th IEEE international
  symposium on}, pages 127--136. IEEE, 2011.

\bibitem{novotny2017learning}
D.~Novotny, D.~Larlus, and A.~Vedaldi.
\newblock Learning 3d object categories by looking around them.
\newblock In {\em Proceedings of the IEEE International Conference on Computer
  Vision}, pages 5218--5227, 2017.

\bibitem{pham2019real}
Q.-H. Pham, B.-S. Hua, T.~Nguyen, and S.-K. Yeung.
\newblock Real-time progressive 3d semantic segmentation for indoor scenes.
\newblock In {\em 2019 IEEE Winter Conference on Applications of Computer
  Vision (WACV)}, pages 1089--1098. IEEE, 2019.

\bibitem{redmon2016you}
J.~Redmon, S.~Divvala, R.~Girshick, and A.~Farhadi.
\newblock You only look once: Unified, real-time object detection.
\newblock In {\em Proceedings of the IEEE conference on computer vision and
  pattern recognition}, pages 779--788, 2016.

\bibitem{riegler2017octnet}
G.~Riegler, A.~Osman~Ulusoy, and A.~Geiger.
\newblock Octnet: Learning deep 3d representations at high resolutions.
\newblock In {\em Proceedings of the IEEE Conference on Computer Vision and
  Pattern Recognition}, pages 3577--3586, 2017.

\bibitem{riegler2017octnetfusion}
G.~Riegler, A.~O. Ulusoy, H.~Bischof, and A.~Geiger.
\newblock Octnetfusion: Learning depth fusion from data.
\newblock In {\em 2017 International Conference on 3D Vision (3DV)}, pages
  57--66. IEEE, 2017.

\bibitem{salas2013slam++}
R.~F. Salas-Moreno, R.~A. Newcombe, H.~Strasdat, P.~H. Kelly, and A.~J.
  Davison.
\newblock Slam++: Simultaneous localisation and mapping at the level of
  objects.
\newblock In {\em Computer Vision and Pattern Recognition (CVPR), 2013 IEEE
  Conference on}, pages 1352--1359. IEEE, 2013.

\bibitem{sun2018pix3d}
X.~Sun, J.~Wu, X.~Zhang, Z.~Zhang, C.~Zhang, T.~Xue, J.~B. Tenenbaum, and W.~T.
  Freeman.
\newblock Pix3d: Dataset and methods for single-image 3d shape modeling.
\newblock In {\em Proceedings of the IEEE Conference on Computer Vision and
  Pattern Recognition}, pages 2974--2983, 2018.

\bibitem{sunderhauf2017meaningful}
N.~S{\"u}nderhauf, T.~T. Pham, Y.~Latif, M.~Milford, and I.~Reid.
\newblock Meaningful maps with object-oriented semantic mapping.
\newblock In {\em 2017 IEEE/RSJ International Conference on Intelligent Robots
  and Systems (IROS)}, pages 5079--5085. IEEE, 2017.

\bibitem{tatarchenko2017octree}
M.~Tatarchenko, A.~Dosovitskiy, and T.~Brox.
\newblock Octree generating networks: Efficient convolutional architectures for
  high-resolution 3d outputs.
\newblock In {\em Proceedings of the IEEE International Conference on Computer
  Vision}, pages 2088--2096, 2017.

\bibitem{whelan2015elasticfusion}
T.~Whelan, S.~Leutenegger, R.~Salas-Moreno, B.~Glocker, and A.~Davison.
\newblock Elasticfusion: Dense slam without a pose graph.
\newblock Robotics: Science and Systems, 2015.

\bibitem{wu2017marrnet}
J.~Wu, Y.~Wang, T.~Xue, X.~Sun, B.~Freeman, and J.~Tenenbaum.
\newblock Marrnet: 3d shape reconstruction via 2.5 d sketches.
\newblock In {\em Advances In Neural Information Processing Systems}, pages
  540--550, 2017.

\bibitem{wu20153d}
Z.~Wu, S.~Song, A.~Khosla, F.~Yu, L.~Zhang, X.~Tang, and J.~Xiao.
\newblock 3d shapenets: A deep representation for volumetric shapes.
\newblock In {\em Proceedings of the IEEE conference on computer vision and
  pattern recognition}, pages 1912--1920, 2015.

\bibitem{yan2016perspective}
X.~Yan, J.~Yang, E.~Yumer, Y.~Guo, and H.~Lee.
\newblock Perspective transformer nets: Learning single-view 3d object
  reconstruction without 3d supervision.
\newblock In {\em Advances in Neural Information Processing Systems}, pages
  1696--1704, 2016.

\bibitem{yang20173d}
B.~Yang, H.~Wen, S.~Wang, R.~Clark, A.~Markham, and N.~Trigoni.
\newblock 3d object reconstruction from a depth view with adversarial learning.
\newblock {\em arXiv preprint arXiv:1708.07969}, 2017.

\bibitem{zhu2017semantic}
R.~Zhu, C.~Wang, C.-H. Lin, Z.~Wang, and S.~Lucey.
\newblock Semantic photometric bundle adjustment on natural sequences.
\newblock {\em arXiv preprint arXiv:1712.00110}, 2017.

\end{thebibliography}
}

\end{document}